\documentclass[conference]{IEEEtran}
\IEEEoverridecommandlockouts
\usepackage{cite}
\usepackage{amsmath,amssymb,amsfonts}
\usepackage{algorithmic}
\usepackage{graphicx}
\usepackage{textcomp}
\usepackage{xcolor}
\def\BibTeX{{\rm B\kern-.05em{\sc i\kern-.025em b}\kern-.08em
    T\kern-.1667em\lower.7ex\hbox{E}\kern-.125emX}}
\begin{document}

\title{CMSBERT-CLR: Context-driven Modality Shifting BERT with Contrastive Learning for linguistic, visual, acoustic Representations}
\author{\IEEEauthorblockN{1\textsuperscript{st} Junghun Kim}
\IEEEauthorblockA{\textit{Department of Artificial Intelligence} \\
\textit{Dongguk University}\\
Seoul, Korea \\
riseup32@dongguk.edu}
\and
\IEEEauthorblockN{2\textsuperscript{nd} Jihie Kim}
\IEEEauthorblockA{\textit{Department of Artificial Intelligence} \\
\textit{Dongguk University}\\
Seoul, Korea \\
jihie.kim@dgu.edu}
}

\maketitle

\begin{abstract}
Multimodal sentiment analysis has become an increasingly popular research area as the demand for multimodal online content is growing. For multimodal sentiment analysis, words can have different meanings depending on the linguistic context and non-verbal information, so it is crucial to understand the meaning of the words accordingly. In addition, the word meanings should be interpreted within the whole utterance context that includes nonverbal information. In this paper, we present a Context-driven Modality Shifting BERT with Contrastive Learning for linguistic, visual, acoustic Representations (CMSBERT-CLR), which incorporates the whole context's non-verbal and verbal information and aligns modalities more effectively through contrastive learning. First, we introduce a Context-driven Modality Shifting (CMS) to incorporate the non-verbal and verbal information within the whole context of the sentence utterance. Then, for improving the alignment of different modalities within a common embedding space, we apply contrastive learning. Furthermore, we use an exponential moving average parameter and label smoothing as optimization strategies, which can make the convergence of the network more stable and increase the flexibility of the alignment. In our experiments, we demonstrate that our approach achieves state-of-the-art results.
\end{abstract}


\section{Introduction}
Within a sentence, words have different meanings depending on the linguistic context. ELMo \cite{b1} used learned functions of the internal states of a deep bidirectional language model to reflect this characteristic. In Multimodal Sentiment Analysis (MSA), non-verbal information is used to help to understand the meaning of words. MAG-BERT \cite{b2} used the informative visual and acoustic factors to adjust the word representations based on the moment's non-verbal information. MulT \cite{b3} combined the non-verbal and verbal information through crossmodal attention by introducing a crossmodal Transformer. However, there are drawbacks to these methods. MAG-BERT took into account non-verbal information in the word shifting process only within the same time step and did not fully exploit the non-verbal information of the entire sentence utterance. This is different from the human inference process, where the word's meaning is understood based on the whole context of the sentence utterance. MulT introduced six crossmodal Transformers to account for all modality pairs. This method attempted to use non-verbal information of the sentence but required a lot of parameters, due to six crossmodal Transformers.

\begin{figure}[t]
\centering
    \includegraphics[width=1 \linewidth]{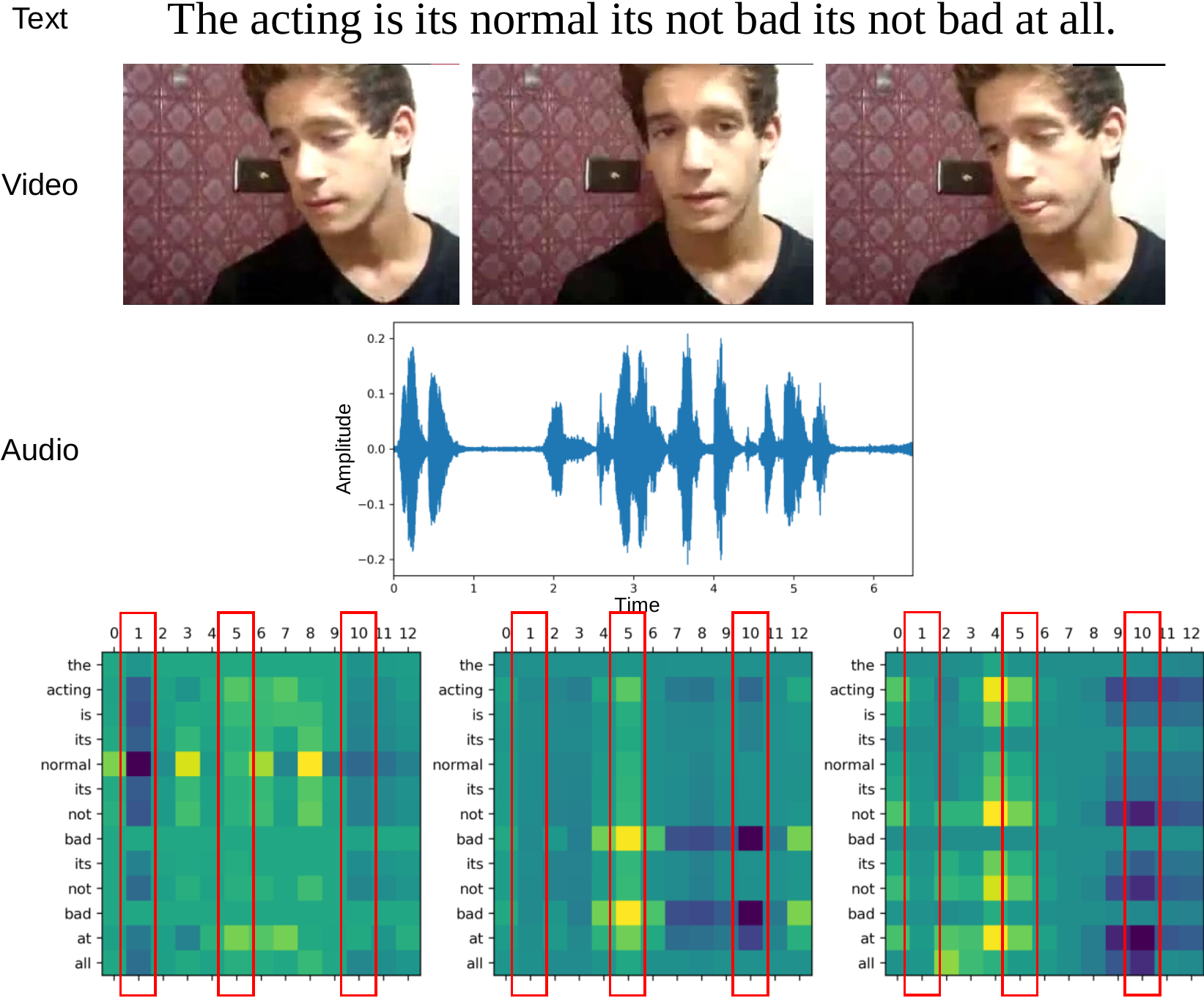}
\caption{The first, second, and third rows are text, video, and audio, respectively. The last row provides an example of Context-driven Modality Shifting (CMS) maps of three different Transformer layers. In the CMS maps, the vertical axis represents the linguistic (verbal) information, and the horizontal one is for visual and acoustic (non-verbal) information.}
\end{figure}

To incorporate the non-verbal and verbal information within the whole context of the sentence utterance more effectively, we introduce Context-driven Modality Shifting (CMS). CMS does not require many parameters because it operates on one Transformer \cite{b4}. CMS can be applied to each Transformer layer in BERT \cite{b5}, enabling each layer’s word representations being shifted based on the entire context. \cite{b6} showed that BERT encodes other properties for each layer, so we conjecture that applying different shifting for each layer is adequate. We call the CMS-applied BERT model CMSBERT. CMSBERT can also be seen as a generalized version of MAG-BERT. In section 3, we compare the two models.

In Fig. 1, we show an example of CMS maps of three different Transformer layers and how non-verbal context information can affect the CMS maps. In the CMS maps, the vertical axis represents the linguistic (verbal) information, and the horizontal one is for visual and acoustic (non-verbal) information. In the CMS maps, the 1st, 5th, and 10th columns of the non-verbal information axis represent the first, second, and third video frames, respectively. The red boxes highlight the mapping. Bright colors mean positive sentiment while darker ones are for negative sentiment. Word representations can be adjusted based on these CMS maps. The word 'normal' in the text has a neutral sensibility. However, in the first CMS map, this word representation is shifted negatively by the `down-looking' gaze processing. Similarly, the word `bad' has a negative sentiment, and in the second CMS map, the word representation is shifted positively or more negatively by the `up-looking' or `down-looking' gaze processing, respectively. The same pattern appears for the word `not' in the third CMS map. These show that even the same word can be embedded differently depending on the non-verbal information of the entire context.

Even when non-verbal information is additionally used, using only CMSBERT may omit important information because it does not directly use intra-modality dynamics of visual and acoustic information. We solve this problem by learning intra-modality and inter-modality dynamics simultaneously. A goal of simultaneously learning intra-modality and inter-modality dynamics is to learn a model that does well in the alignment among the modalities based on expressive representations within each modality. To achieve the goal, MISA \cite{b7} introduced a modality-invariant encoder that projects each modality to a common space and used Central Moment Discrepancy (CMD) \cite{b8} as a similarity loss to align cross-modal representations within the common space. However, we find that while MISA mapped the representations to the common embedding space, non-linguistic modalities are not well identified. To mitigate this problem, we adopt contrastive learning that can optimize the alignment of all the modalities in the common space.

Motivated by \cite{b9} that optimizes bidirectional contrastive loss between visual and linguistic modalities, we introduce a pairwise contrastive loss among linguistic, visual, and acoustic modalities and optimize the pairwise contrastive loss along with the Mean Squared Error (MSE) loss. In the optimization process, we use two techniques to improve the convergence stability and the alignment, as described below. Because the MSE loss and the pairwise contrastive loss differ in the convergence speed and scale, optimizing the pairwise contrastive loss along with MSE loss is a challenge. To handle this problem, we use an exponential moving average parameter to adjust the scale and make the convergence of the network more stable. The same positive or negative sentiment can be expressed in various non-verbal expressions. Therefore, defining positive and negative samples with individual instances in contrastive learning is not appropriate. For this problem, we adopt label smoothing in the alignment process of contrastive learning.

With all of these, we present a Context-driven Modality Shifting BERT with Contrastive Learning for linguistic, visual, acoustic Representations (CMSBERT-CLR), which combines CMSBERT with contrastive learning and stable alignment methods described above. In the experiments, we study the commonly used CMU-MOSI \cite{b10} and CMU-MOSEI \cite{b11} datasets, and we demonstrate that our approach achieves state-of-the-art results.

Our contributions can be summarized as follows:
\begin{itemize}
	\item We present a Context-driven Modality Shifting BERT (CMSBERT) that can attend to non-verbal information of the entire context and does not require many parameters.
	\item To help the alignment among modalities in the common space, we present a Context-driven Modality Shifting BERT with Contrastive Learning for linguistic, visual, acoustic Representations (CMSBERT-CLR), which combines CMSBERT with contrastive learning. For contrastive learning, we use an exponential moving average and label smoothing as optimization strategies.
	\item We demonstrate that our approach achieves state-of-the-art results.
\end{itemize} 

\section{Related Works}
\subsection{Multimodal Sentiment Analysis}
Prior works in multimodal sentiment analysis have learned representations of multiple modalities via various approaches. Early fusion methods that concatenate multimodal data at the input level, such as \cite{b10, b12}, can learn inter-modality dynamics, but they have limitations in learning intra-modality dynamics. Late fusion methods that integrate different modalities at the prediction level, such as \cite{b13, b14}, focus on modeling intra-modality dynamics rather than inter-modality dynamics. Some studies introduced additional modules to model both intra-modality and inter-modality dynamics. MISA \cite{b7} introduced a modality-invariant encoder that projects each modality to a common space and used Central Moment Discrepancy (CMD) \cite{b8} as a similarity loss to align cross-modal representations within the common space. Self-MM \cite{b14} presented a label generation module based on self-supervised learning to acquire independent unimodal supervisions. There were also studies focused on changes in word representations. RAVEN \cite{b15}, and MAG-BERT \cite{b2} made use of non-verbal information to shift the word representations within the same time step. MulT \cite{b3} combined the non-verbal and verbal information through crossmodal attention by introducing six crossmodal Transformers, which require a lot of parameters.

In this paper, we propose a novel Context-driven Modality Shifting (CMS) that takes into account of the entire context of the sentence utterance and does not require many parameters. CMS can be applied individually to all the layers.

\begin{figure*}[t]
\centering
    \includegraphics[width=0.75 \linewidth]{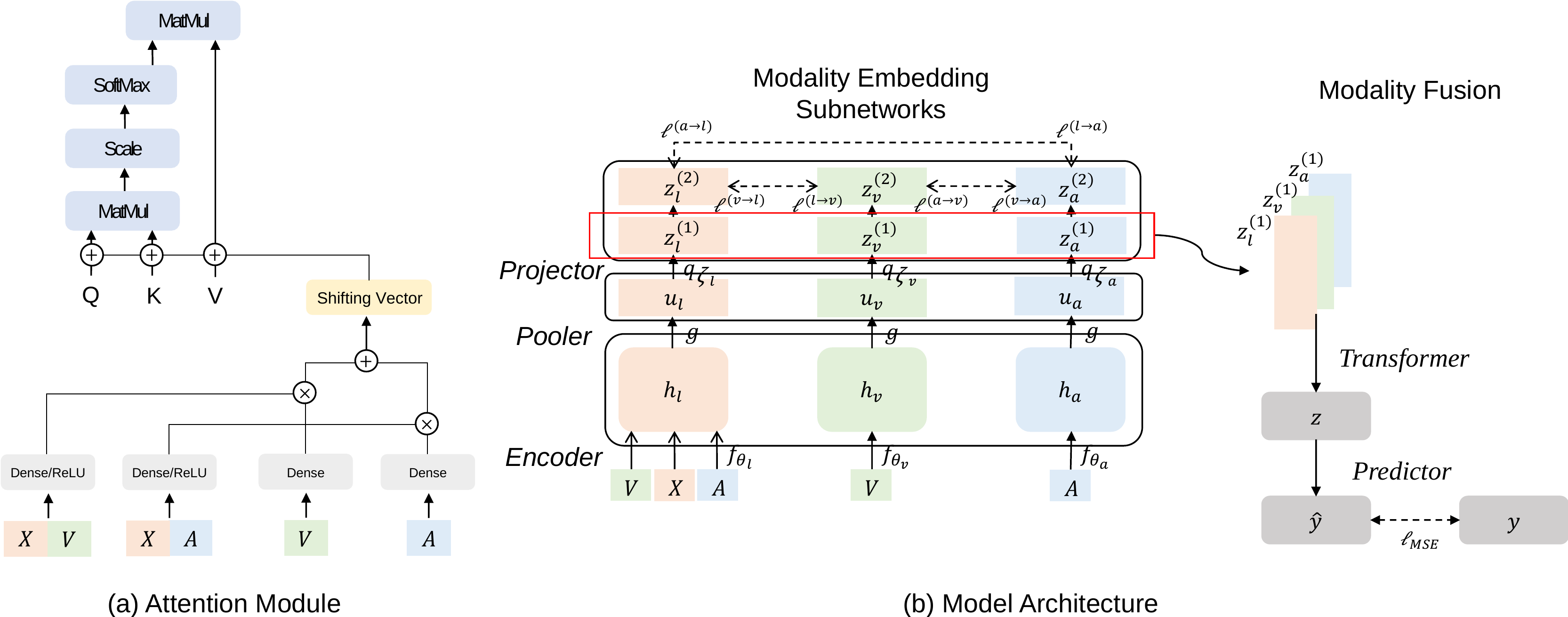}
\caption{(a) is a self-attention module in CMSBERT. (b) is a proposed CMSBERT-CLR architecture. CMSBERT-CLR minimizes an MSE loss and a pairwise contrastive loss simultaneously. Modality embedding subnetworks are comprised of three stages: an encoder $f_{\theta_m}$, a pooler $g$, and a projector $q_{\zeta_m}$, where $\theta$ and $\zeta$ are model parameters and $m$ denotes linguistic ($l$), visual ($v$) and acoustic ($a$) modalities.}
\end{figure*}

\subsection{Contrastive Learning}
Contrastive learning is a self-supervised method that does not require specialized architecture and can significantly improve performance from unlabeled data. Contrastive approaches learn to bring positive pairs closer, and keep negative pairs apart. Several tasks (object detection, instance segmentation, keypoint detection) in computer vision \cite{b16, b17, b18, b19, b20, b21, b22, b23} have shown significant performance improvement with a simple model architecture along with contrastive learning. Experiments for natural language processing \cite{b24, b25, b26, b27} and speech recognition \cite{b28} tasks also have demonstrated the effectiveness of contrastive learning. Contrastive approaches can be used not only for self-supervised learning but also for supervised learning. In classification tasks \cite{b29, b30}, contrastive loss allows the same classes to be aligned closer in the embedding space, making the decision boundaries clearer. In addition, contrastive learning is available in multimodality \cite{b9, b31, b32} as well as unimodality. It improves representation learning by maximizing the agreement between true versus random pairs through bidirectional contrastive loss between each pair. In particular, \cite{b33} introduced cross-modal agreement in contrastive learning to treat similar instances as positive samples rather than defining positive and negative samples from individual instances.

In this work, we adopt contrastive learning for improving the alignment of different modalities. Furthermore, we use the exponential moving average parameter to make the convergence stable and label smoothing to calibrate the similar instances' labels for sentiment learning.

\section{Approach}
We first introduce a novel Context-driven Modality Shifting (CMS) to incorporate the non-verbal and verbal information within the entire context of the sentence utterance. Then, we describe modality-specific encoder models for capturing intra-modality dynamics and a Transformer-based fusion method for integrating inter-modality dynamics. Finally, we adopt contrastive learning and optimization strategies such as the exponential moving average parameter and the label smoothing to align cross-modal representations more effectively. The overall model architecture is illustrated in Fig. 2.

\subsection{Context-driven Modality Shifting}
MAG-BERT \cite{b2} proposed a method of shifting the word representations using non-verbal information. However, MAG-BERT took into account non-verbal information only within the same time step in the word shifting process and did not fully exploit the non-verbal information of the entire sentence utterance. This is different from the human inference process, where the word's meaning is understood based on the whole context of the sentence utterance. MulT \cite{b3} combined the non-verbal and verbal information of the entire sentence utterance through crossmodal attention but required a lot of parameters because of six crossmodal Transformers. To mitigate this weakness, we introduce a novel Context-driven Modality Shifting (CMS). CMS attends to non-verbal information within the entire context of the sentence utterance and does not require many parameters. CMS is applied to each Transformer layer in BERT, enabling different shifting for each layer's word representations. \cite{b6} showed that BERT encodes other properties for each layer, so we conjecture that applying different shifting for each layer is adequate. CMS is involved in the self-attention process of Transformer and has a similar structure to the relation-aware self-attention in \cite{b34}. The word representations shifting process through non-verbal information is as follows:
$$
g_{ij}^{v}=\sigma(W_{gv}[X_{i};V_{j}]+b_{v}) \in \mathbb{R}, \eqno{(1)}
$$

$$
g_{ij}^{a}=\sigma(W_{ga}[X_{i};A_{j}]+b_{a}) \in \mathbb{R}, \eqno{(2)}
$$

$$
a_{ij}=g_{ij}^{v}\cdot(W_{v}V_{j})+g_{ij}^{a}\cdot(W_{a}A_{j}) \in \mathbb{R}^{d}, \eqno{(3)}
$$
where $X_{i}$ is a $i$-th word representation and $V_{j}$ and $A_{j}$ are $j$-th information for visual and acoustic. $W_{gv}$, $W_{ga}$, $W_{v}$ and $W_{a}$ are weight matrices and $b_{v}$ and $b_{a}$ are scalar biases. $\sigma$ is non-linear activation function and $[\cdot;\cdot]$ denotes concatenation. Eq. (3) means that the $i$-th word representation is affected by the $j$-th visual and acoustic information.

We average the vectors so that the $i$-th word representation considers the entire context's non-verbal information. The self-attention is converted as follows and individually applied to each Transformer layer:
$$
a_i=\frac{1}{N}\sum_{j=1}^{N}a_{ij} \in \mathbb{R}^{d}, \eqno{(4)}
$$

$$
e_{ij}=\frac{(X_{i}W^{Q}+a_{i})(X_{j}W^{K}+a_{j})^{T}}{\sqrt{d}} \in \mathbb{R}, \eqno{(5)}
$$

$$
\alpha_{ij}=\frac{\exp{e_{ij}}}{\sum_{k=1}^N\exp{e_{ik}}} \in \mathbb{R}, \eqno{(6)}
$$

$$
z_{i}=\sum_{j=1}^N\alpha_{ij}(X_{j}W^{v}+a_{j}) \in \mathbb{R}^{d}. \eqno{(7)}
$$

We call the CMS-applied BERT model CMSBERT. CMSBERT can be seen as a generalized version of MAG-BERT. The shifting operation of MAG-BERT is applied only once in the $l$-th layer, and general self-attention is used in other layers. If we apply CMS only once and $a_{11}=\cdot\cdot\cdot=a_{1N}=H_{1}^{'}$, $a_{21}=\cdot\cdot\cdot=a_{2N}=H_{2}^{'}$ and $a_{N1}=\cdot\cdot\cdot=a_{NN}=H_{N}^{'}$, then CMSBERT becomes equivalent to MAG-BERT.

\subsection{Modality Embedding Subnetworks}
All our modality embedding subnetworks are comprised of three stages: an encoder $f_{\theta_m}$, a pooler $g$, and a projector $q_{\zeta_m}$, where $\theta_m$ and $\zeta_m$ are model parameters and $m$ denotes linguistic ($l$), visual ($v$) and acoustic ($a$) modalities. First, for each modality $m \in \{l, v, a\}$, we use the encoder to map $N$ sequence $X, V, A \in \mathbb{R}^{N \times d_m}$ to an equal sized representations $h_m \in \mathbb{R}^{N \times d}$. Then, we pool $h_m$ to a pooled vectors $u_m \in \mathbb{R}^{d}$. The pooled vectors $u_m$ is projected to a final modality representations $z_m \in \mathbb{R}^{d}$.

\textbf{Linguistic Embedding Subnetwork:} Following recent approaches \cite{b2, b7, b14}, we utilize the pre-trained BERT's embedding as the feature extractor for word tokens. Given a length $N$ word tokens $W = [W_1, W_2, ..., W_N] \in \mathbb{R}^{N}$, we input $W$ to the input embedder which outputs $X = [X_1, X_2, ... ,X_N] \in \mathbb{R}^{N \times d_l}$. The linguistic feature dimensions, $d_l$, are 768 for MOSI and MOSEI, which are the dataset we use for our evaluation, as described below in the experiment section.

Given the $X$, $V$ and $A$, we use the CMSBERT as the linguistic encoder ($f_{\theta_l}$).
$$
h_l = \mathrm{CMSBERT}(X, V, A; {\theta_l}). \eqno{(8)}
$$
To get the pooled linguistic vector $u_l$, we pool $h_l$ using mean pooler $g$.
$$
u_l = g(h_l). \eqno{(9)}
$$
Mean pooler $g$ does not require any learning parameter. The pooled linguistic vector $u_l$ is then transformed with the non-linear projector $q_{\zeta_l}$, yielding $z_l$ that is used for contrastive learning.
$$
z_l = q(u_l; {\zeta_l}), \eqno{(10)}
$$
where the projector is consists of a two-layer projection head.

\textbf{Visual Embedding Subnetwork:} Both MOSI and MOSEI use Facet to extract facial expression features, which include facial action units and face pose based on the Facial Action Coding System (FACS) \cite{b35}. This process is repeated for each sampled frame within the utterance video sequence, which outputs $N$ length visual feature $V = [V_1, V_2, ..., V_N] \in \mathbb{R}^{N \times d_v}$. The visual feature dimensions, $d_v$, are 47 for MOSI, 35 for MOSEI.

Given the $V$, we use a Long Short-Term Memory (LSTM) \cite{b36} as the visual encoder ($f_{\theta_v}$).
$$
h_v = \mathrm{LSTM}(V; {\theta_v}). \eqno{(11)}
$$
To get the pooled visual vector $u_v$, we pool the $h_v$ using mean pooler $g$.
$$
u_v = g(h_v). \eqno{(12)}
$$
As before, the mean pooler $g$ does not require any learning parameter. The pooled visual vector $u_v$ is then transformed with the non-linear projector $q_{\zeta_v}$, yielding $z_v$ that is used for contrastive learning.
$$
z_v = q(u_v; {\zeta_v}), \eqno{(13)}
$$
where the projector is consists of the two-layer projection head.

\textbf{Acoustic Embedding Subnetwork:} Both MOSI and MOSEI use COVAREP \cite{b37} to extract various low-level statistical audio features, which include 12 Mel-frequency cepstral coefficients, pitch, Voiced/Unvoiced segmenting features (VUV) \cite{b38}, glottal source parameters \cite{b39}, and other features related to emotions and tone of speech. This process is repeated for each sampled frame within the utterance sequence, which outputs $N$ length acoustic feature $A = [A_1, A_2, ..., A_N] \in \mathbb{R}^{N \times d_a}$. The acoustic feature dimensions, $d_a$, are 74 for MOSI and MOSEI.

The architecture of the acoustic embedding subnetwork is the same as the visual embedding subnetwork.

\subsection{Modality Fusion}
After projecting the modalities into their respective representations, we fuse them into a joint vector for downstream prediction. To capture the inter-modality dynamics, following \cite{b7}'s study, we use a Transformer \cite{b4} architecture as a fusion method. We concatenate the representations ($z_m^{(1)}$) from the middle layer of each projection head, as \cite{b20} showed that fine-tuning from the middle layer leads to better downstream task performance.
$$
z^{'} = \mathrm{Concat}(z_l^{(1)}, z_v^{(1)}, z_a^{(1)}) \in \mathbb{R}^{3 \times d}. \eqno{(14)}
$$
Then the concatenated vector $z^{'}$ is used as input into Transformer and predicted with the non-linear predictor.
$$
z = \mathrm{Reshape}(\mathrm{Transformer}(z^{'})) \in \mathbb{R}^{3d}, \eqno{(15)}
$$
$$
\widehat{y} = W_2\mathrm{ReLU}(W_1z+b_1)+b_2 \in \mathbb{R}, \eqno{(16)}
$$
where $W_1$ and $W_2$ are weight matrices, $b_1$ and $b_2$ are bias vectors.

\subsection{Contrastive Learning}

\textbf{Pairwise Contrastive Loss.} Motivated by ConVIRT \cite{b9} that optimizes bidirectional contrastive loss between visual and linguistic modalities, we extend the bidirectional contrastive loss to the pairwise contrastive loss among linguistic, visual, and acoustic modalities. The following describes how we handle linguistic and visual pairs as an example. We sample a minibatch of $n$ input pairs ($X$, $V$) and obtain their final modality representations ($z_l$, $z_v$). We use (${z_l}_i$, ${z_v}_i$) to denote the $i$-th pair in minibatch. This pair has two loss functions. The first one is a visual-to-linguistic contrastive loss for the $i$-th pair:
$$
\ell_i^{(v \rightarrow l)}=-\mathrm{log}\frac{\mathrm{exp}(\langle {z_v}_i, {z_l}_i \rangle/\tau)}{\sum_{k=1}^n\mathrm{exp}(\langle {z_v}_i, {z_l}_k \rangle/\tau)}, \eqno{(17)}
$$
where $\langle \cdot, \cdot \rangle$ denotes the cosine similarity, and $\tau \in \mathbb{R}^{+}$ denotes a temperature parameter. Furthermore, we use the asymmetric contrastive loss for each linguistic-to-visual modality pair. We thus define a linguistic-to-visual loss for the $i$-th pair as follow:
$$
\ell_i^{(l \rightarrow v)}=-\mathrm{log}\frac{\mathrm{exp}(\langle {z_l}_i, {z_v}_i \rangle/\tau)}{\sum_{k=1}^n\mathrm{exp}(\langle {z_l}_i, {z_v}_k \rangle/\tau)}. \eqno{(18)}
$$
This process applies equally to linguistic and acoustic pairs and visual and acoustic pairs. In other words, we have a total of six contrastive losses. We thus define the pairwise contrastive loss as follows:
$$
\ell_{Con}=\sum_\mathcal{D}\ell^{(\mathcal{D})}, \eqno{(19)}
$$
where $\mathcal{D}=\{l \rightarrow v, v \rightarrow l, l \rightarrow a, a \rightarrow l, v \rightarrow a, a \rightarrow v\}$.

\textbf{Exponential Moving Average Parameter.} MSE loss and pairwise contrastive loss differ in convergence speed and scale, which can make the learning unstable. To adjust the scale between two losses and make the convergence of the network more stable, we use an exponential moving average parameter $\nu$ and define the final loss as follows:
$$
\ell_{total}=\ell_{MSE}+\nu\ell_{Con}, \eqno{(20)}
$$
$$
\nu=\alpha \nu+(1-\alpha)\widehat{\nu}, \eqno{(21)}
$$
$$
\widehat{\nu}=\frac{\ell_{MSE}^{(i)}}{\ell_{Con}^{(i)}}, \eqno{(22)}
$$
where $\alpha$ is a momentum coefficient and $i$ is $i$-th validation epoch.

\textbf{Label Smoothing.}
Even with the same verbal expression, the sentiment can be different depending on the non-verbal elements. Also, there are various non-verbal ways of expressing the same sentiment. Therefore, defining positive and negative samples with individual instances in contrastive learning is not appropriate. To deal with this problem, we use label smoothing in contrastive learning, which can align verbals and non-verbals of similar sentiment flexibly. We define the label in contrastive loss of $i$ and $j$ pair as follows:
$$
{y_{con}}_{ij}=\begin{cases}\beta & ,y_{ii} 
\\ \frac{1-\beta}{|S|} & ,S 
\\ 0 & ,otherwise\end{cases}, \eqno{(23)}
$$
where $\beta$ is a smoothing coefficient for the true pairs and $S=\{y_{ij}|sign(y_{i})=sign(y_{j}), |y_{i}-y_{j}|\leq \delta \}$, i.e., a set with the same sign sentiment scores and the sentiment scores' difference less than $\delta$. $\delta$ is a threshold parameter. In this paper, we set $\delta$ to 0.1 through experiments.

\section{Experiments}
\subsection{Datasets}

We use two public multimodal sentiment analysis datasets, CMU-MOSI and CMU-MOSEI, which are widely used in MSA evaluations. CMU-MOSI contains 2,199 video segments taken from 93 Youtube movie review videos. CMU-MOSEI contains 23,453 annotated video segments taken from 5,000 videos, 1,000 distinct speakers, and 250 different topics. CMU-MOSEI has more utterances and greater variety in samples, speakers, and topics than CMU-MOSI. In both CMU-MOSI and CMU-MOSEI, human annotators label each sample with a sentiment score from -3 to 3.

\begin{table*}[t]
\caption{
Comparison results with the state-of-the-art methods on CMU-MOSI and CMU-MOSEI.
}
\begin{center}
\begin{tabular}{c|cccc|cccc}
\hline
 & & & \textbf{MOSI} & & & & \textbf{MOSEI} & \\
\hline
Method & MAE $\downarrow$ & Corr $\uparrow$ & Acc-2 $\uparrow$ & F1-Score $\uparrow$ & MAE $\downarrow$ & Corr $\uparrow$ & Acc-2 $\uparrow$ & F1-Score $\uparrow$ \\
\hline
TFN (B) \cite{b40} & 0.901 & 0.698 & -/80.80 & -/80.70 & 0.593 & 0.700 & -/82.50 & -/82.10 \\
MFN (G) \cite{b41} & 0.965 & 0.632 & 77.40/- & 77.30/- & - & - & 76.00/- & 76.00/- \\
RAVEN (G) \cite{b15} & 0.915 & 0.691 & 78.00/- & 76.60/- & 0.614 & 0.662 & 79.10/- & 79.50/- \\
MFM (B) \cite{b42} & 0.877 & 0.706 & -/81.70 & -/81.60 & 0.568 & 0.717 & -/84.40 & -/84.30 \\
MulT (B) \cite{b3} & 0.861 & 0.711 & 81.50/84.10 & 80.60/83.90 & 0.580 & 0.703 & -/82.50 & -/82.30 \\
MISA (B) \cite{b7} & 0.804 & 0.764 & 80.79/82.10 & 80.77/82.03 & 0.568 & 0.724 & 82.59/84.23 & 82.67/83.97 \\
MAG-BERT (B) \cite{b2} & 0.731 & 0.789 & 82.54/84.30 & 82.59/84.30 & 0.539 & 0.753 & 83.79/85.23 & 83.74/85.08 \\
Self-MM (B) \cite{b14} & 0.713 & 0.798 & 84.00/85.98 & 84.42/85.95 & 0.530 & 0.765 & 82.81/85.17 & 82.53/85.30 \\
\hline
CMSBERT (B) & 0.711 & 0.795 & 83.65/85.65 & 83.57/85.62 & 0.529 & 0.769 & 83.76/86.24 & 83.78/86.18 \\
CMSBERT-CLR (B) & \textbf{0.708} & \textbf{0.802} & \textbf{84.82}/\textbf{86.56} & \textbf{84.73}/\textbf{86.48} & \textbf{0.523} & \textbf{0.776} & \textbf{84.39}/\textbf{86.69} & \textbf{84.27}/\textbf{86.63} \\
\hline
\end{tabular}
\end{center}
\end{table*}

\subsection{Baselines}
We compare the performance of our approach with a variety of state-of-the-art methods for multimodal sentiment analysis. \\
\textbf{TFN (Tensor Fusion Network)} \cite{b40} used a tensor fusion to capture uni-, bi-, and tri-modal interactions. \\
\textbf{MFN (Memory Fusion Network)} \cite{b41} modeled each modality separately and synchronized among them using multi-view gated memory. \\
\textbf{RAVEN (Recurrent Attended Variation Embedding Network)} \cite{b15} modeled the fine-grained structure of non-verbal subword sequences and shifted dynamically word representations based on non-verbal information. \\
\textbf{MFM (Multimodal Factorization Model)} \cite{b42} learned generative representations and discriminative representations simultaneously to learn the modality-specific generative features and classification. \\
\textbf{MulT (Multimodal Transformer for Unaligned Multimodal Language Sequence)} \cite{b3} learned representations directly from unaligned multimodal streams using the cross-modal attention module. \\
\textbf{MISA (Modality-Invariant and -Specific Representations for Multimodal Sentiment Analysis)} \cite{b7} projected each modality to two distinct subspaces, modality-invariant and modality-specific, to provide a holistic view of the multimodal data. \\
\textbf{MAG-BERT (Multimodal Adaptation Gate for Bert)} \cite{b2} attached Multimodal Adaptation Gate (MAG) to BERT to accept multimodal non-verbal information during fine-tuning. \\ 
\textbf{Self-MM (Self-Supervised Multi-task Multimodal sentiment analysis network)} \cite{b14} designed a label generation module based on the self-supervised learning strategy to acquire independent unimodal supervisions.

\subsection{Experimental Design}
\textbf{Experimental Details.} All the models in this paper are trained using AdamW optimizer \cite{b43} with a learning rate of 1e-4 and apply weight decay of 0.1. We train for 500 epochs with a linear learning rate decay and a warmup period of 50 epochs. A weight decay of 1e-3 is used. We use a batch size of 128 in 8 RTX-2080Ti GPU. We use temperature parameter $\tau$ of \{0.07, 0.1, 0.2\}, $\alpha$ of \{0.9, 0.99, 0.999\} and $\beta$ of \{0.7, 0.8, 0.9\} for training each model. All the models use the designated validation set of CMU-MOSI and CMU-MOSEI for finding the best hyperparameters. In this paper, we use $\tau$ of 0.1, $\alpha$ of 0.9, and $\beta$ of 0.9 in both CMU-MOSI and CMU-MOSEI.

\textbf{Evaluation Metrics.} Following the previous works \cite{b7, b14}, we perform two different evaluation tasks on CMU-MOSI and CMU-MOSEI: classification and regression. We report the weighted F1 score (F1-Score) and binary classification accuracy (Acc-2) for classification. We report Mean Absolute Error (MAE) and Pearson Correlation (Corr) for regression.

\begin{figure}[t]
\centering
    \includegraphics[width=1 \linewidth]{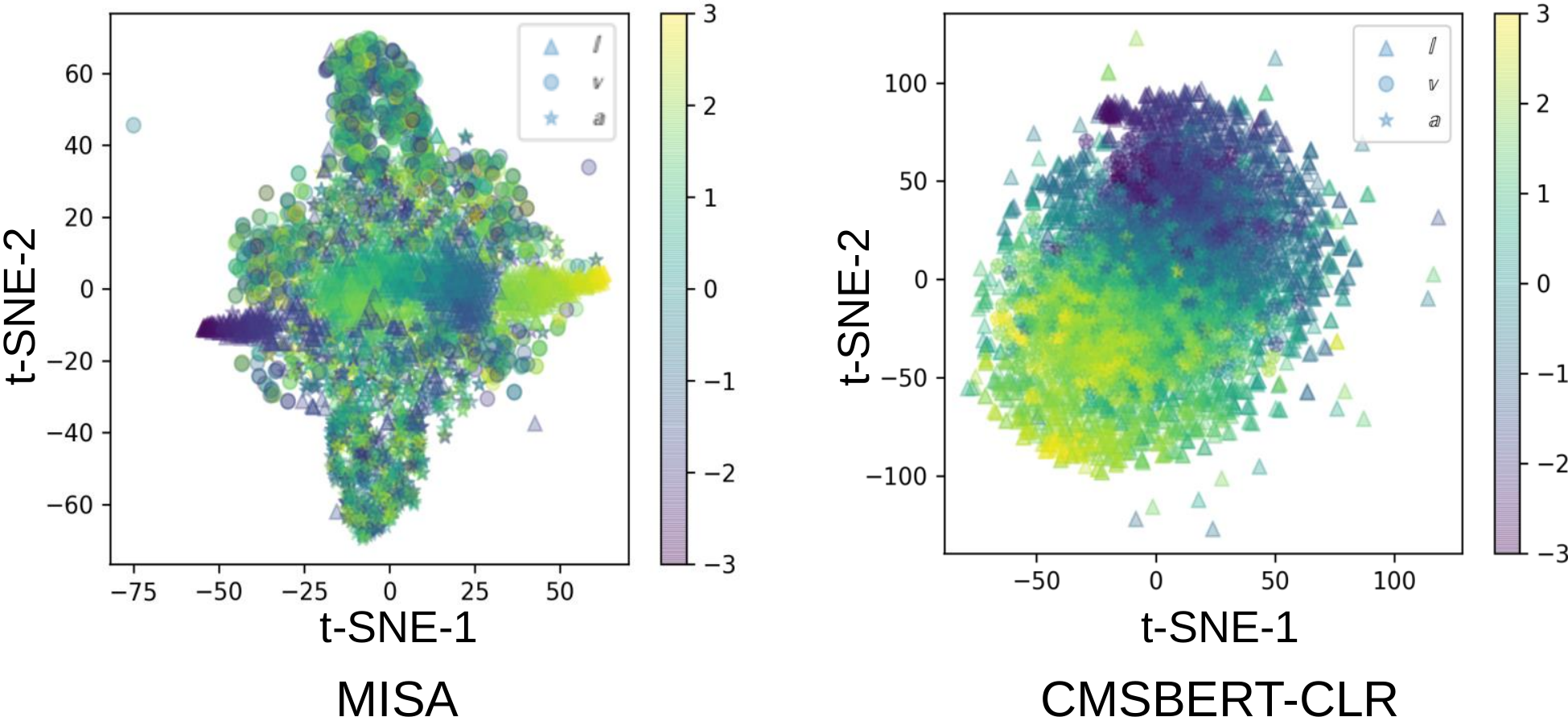}
\caption{t-SNE \cite{b44} visualization of the linguistic, visual, acoustic representations. Triangle is the linguistic modality, the circle is the visual modality, and the star is the acoustic modality. Colors represent sentiment scores from -3 to 3.}
\end{figure}

\section{Results}
\subsection{Comparison with State-of-the-art Methods}
Table I presents a comparison with the state-of-the-art methods over CMU-MOSI and CMU-MOSEI. (B) and (G) mean the use of linguistic features based on BERT and Glove, respectively. In Acc-2 and F1-Score, the left of "/" shows the score with "negative/non-negative" (not excluding zero), and the right is the one with "negative/positive" (excluding zero). In our experiments, CMSBERT achieves comparable performance to state-of-the-art methods. CMSBERT-CLR achieves superior performance over state-of-the-art methods in all the evaluation metrics.

Fig. 3 compares t-SNE \cite{b44} visualization of the linguistic, visual, acoustic representations of MISA and CMSBERT-CLR. MISA that uses modality-invariant encoder and CMD loss maps the representations to the common embedding space. However, sentiments are not distinguished except for the linguistic modality in the common embedding space. In contrast to MISA, in the common embedding space of CMSBERT-CLR, the sentiments of all the modalities are more clearly distinguished. This indicates that contrastive learning can effectively align representations among the modalities.

\begin{table}[t]
\caption{
Comparison results of BERT variants.
}
\begin{center}
\begin{tabular}{c|cccc}
\hline
Method & MAE $\downarrow$ & Corr $\uparrow$ & Acc-2 $\uparrow$ & F1-Score $\uparrow$ \\
\hline
BERT & 0.766 & 0.769 & 82.29/84.12 & 82.18/84.18 \\
MAG-BERT & 0.740 & 0.778 & 83.07/84.58 & 82.97/84.52 \\
CMSBERT/1 & 0.726 & 0.790 & 83.50/85.12 & 83.42/85.14 \\
CMSBERT & \textbf{0.711} & \textbf{0.795} & \textbf{83.65}/\textbf{85.65} & \textbf{83.57}/\textbf{85.62} \\
\hline
\end{tabular}
\end{center}
\end{table}

\begin{table}[t]
\caption{
Comparison results according to contrastive learning and optimization strategies.
}
\begin{center}
\begin{tabular}{c|cccc}
\hline
Method & MAE $\downarrow$ & Corr $\uparrow$ & Acc-2 $\uparrow$ & F1-Score $\uparrow$ \\
\hline
CMSBERT & 0.711 & 0.795 & 83.65/85.65 & 83.57/85.62 \\
\hline
CMSBERT-CLR & \textbf{0.708} & \textbf{0.802} & \textbf{84.82}/\textbf{86.56} & \textbf{84.73}/\textbf{86.48} \\
w/o EMA & 0.709 & 0.798 & 84.53/86.42 & 84.46/86.37 \\
w/o LS & 0.721 & 0.790 & 83.50/85.19 & 83.42/85.19 \\
w/o EMA, LS & 0.730 & 0.784 & 83.36/85.12 & 83.31/85.14 \\
\hline
\end{tabular}
\end{center}
\end{table}

\subsection{Ablation Study}
\textbf{Importance of the architecture of BERT.} Table II shows the performance of BERT variants reproduced under the same conditions for CMU-MOSI. CMSBERT/1 denotes the model with CMS applied to a single layer (first layer), similar to MAG-BERT. Compared to the basic BERT that uses only linguistic modality, MAG-BERT with non-verbal information of the same time step shows improved results. CMSBERT/1 shows performance improvement over MAG-BERT, which implies that non-verbal information for the entire context helps understand the sentiment. Finally, it shows the best performance when CMS is applied separately to all the layers. This is because BERT encodes different properties for each layer.

\textbf{Importance of contrastive learning and optimization strategies.} We design ablation experiments to investigate the effectiveness of contrastive learning and optimization strategies. The results are shown in Table III, where EMA is the exponential moving average parameter, and LS is label smoothing. When CLR is combined with CMSBERT, the performance is significantly increased. This implies that it is crucial to directly learn intra-modality dynamics and align representations among the modalities through contrastive learning. Eliminating the exponential moving average parameter causes performance degradation. Since the pairwise contrastive loss is about four times larger than the MSE loss, adjusting the scale through the exponential moving average helps stable convergence. When label smoothing is removed, the performance is poorer than CMSBERT. Removing label smoothing means that only one is a positive sample, and the rest is all negative samples in contrastive learning. In this case, if there are similar sentiments in the batch, representations cannot be adequately learned. Therefore, the flexible alignment of similar sentiment verbals and non-verbals seems critical in MSA. The performance degradation is most significant when both exponential moving average parameter and label smoothing are removed. In conclusion, combining CMSBERT with a modality fusion method based on contrastive learning and two optimization strategies presents the best performance.

\subsection{Experiment for Emotion Recognition}
In order to evaluate the effectiveness of our model for other tasks, we evaluated our model in emotion recognition. We compared ours with the recent emotion recognition results using the IEMOCAP dataset \cite{b45}. We perform 10-fold cross-validation using only linguistic and acoustic modalities because the baseline methods do not use visual information. The Weighted Accuracy (WA) and Unweighted Accuracy (UA) results are shown in Table IV. Our method achieves state-of-the-art performance in both metrics.

\begin{table}[t]
\caption{
Results of emotion recognition experiment.
}
\begin{center}
\begin{tabular}{c|cc}
\hline
Method & WA $\uparrow$ & UA $\uparrow$ \\
\hline
MHA-3 \cite{b46} & 74.0 & 75.3 \\
F-III \cite{b47} & 79.2 & 80.5 \\
AugPauseRoBERTa \cite{b48} & 72.5 & - \\
CMSBERT-CLR & \textbf{80.8} & \textbf{81.3} \\
\hline
\end{tabular}
\end{center}
\end{table}

\section{Conclusion}
In this paper, we present a Context-driven Modality Shifting BERT with Contrastive Learning for linguistic, visual, acoustic Representations (CMSBERT-CLR), which incorporates the whole context's non-verbal and verbal information into the model and aligns modalities effectively through contrastive learning. CMSBERT can attend to non-verbal information of the entire context without requiring many parameters because only one Transformer is needed. Also, for contrastive learning, we use an exponential moving average parameter to make the convergence stable and label smoothing to calibrate the similar instances' labels. In our experiments, we demonstrate that our approach achieves state-of-the-art results. Although the current work focuses on multimodal sentiment analysis, we plan to expand this approach in other multimodal tasks in the future.

\section{Acknowledgement}
This research was supported by the MSIT(Ministry of Science and ICT), Korea, under the ITRC (Information Technology Research Center) support program (IITP-2022-2020-0-01789) and under the High-Potential Individuals Global Training Program (2021-0-01549) supervised by the IITP(Institute for Information \& Communications Technology Planning \& Evaluation).

\end{document}